\documentclass[letterpaper]{article} 
\usepackage{aaai2026}  
\usepackage{times}  
\usepackage{helvet}  
\usepackage{courier}  
\usepackage[hyphens]{url}  
\usepackage{graphicx} 
\usepackage{natbib}  
\usepackage{caption} 
\usepackage{algorithm}
\usepackage{algorithmic}
\usepackage{booktabs}
\usepackage{amsmath}
\usepackage{amsfonts}

\usepackage[most]{tcolorbox}
\usepackage{enumitem}
\usepackage{lipsum}
\usepackage{threeparttable}
\usepackage{xcolor}
\usepackage{colortbl}
\usepackage{makecell}
\definecolor{mygreen}{rgb}{0,0.5,0}

\usepackage{newfloat}
\usepackage{listings}
\DeclareCaptionStyle{ruled}{labelfont=normalfont,labelsep=colon,strut=off} 
\floatstyle{ruled}
\newfloat{listing}{tb}{lst}{}
\floatname{listing}{Listing}
\title{anyECG-chat: A Generalist ECG-MLLM for Flexible ECG Input and Multi-Task Understanding}
\author{
    Haitao Li\textsuperscript{\rm 1,2}, 
    Ziyu Li\textsuperscript{\rm 1}, 
    Yiheng Mao\textsuperscript{\rm 1}, 
    Ziyi Liu\textsuperscript{\rm 3}, 
    Zhoujian Sun\textsuperscript{\rm 4}, 
    Zhengxing Huang\textsuperscript{\rm 1}\\
}
\affiliations{
    \textsuperscript{\rm 1}Zhejiang University\\
    \textsuperscript{\rm 2}Shanghai Innovation Institute\\
    \textsuperscript{\rm 3}Transtek Medical Electronics Co., Ltd.\\
    \textsuperscript{\rm 4}Ant Group\\
    lihaitao@zju.edu.cn, zhengxinghuang@zju.edu.cn
}

\usepackage{bibentry}

\begin{document}

\maketitle
\begin{abstract}
The advent of multimodal large language models (MLLMs) has sparked interest in their application to electrocardiogram (ECG) analysis. 
However, existing ECG-focused MLLMs primarily focus on report generation tasks, often limited to single 12-lead, short-duration (10s) ECG inputs, thereby underutilizing the potential of MLLMs. To this end, we aim to develop a MLLM for ECG analysis that supports a broader range of tasks and more flexible ECG inputs. 
However, existing ECG-QA datasets are often monotonous. To address this gap, we first constructed the anyECG dataset, which encompasses a wide variety of tasks, including report generation, abnormal waveform localization, and open-ended question answering. In addition to standard hospital ECGs, we introduced long-duration reduced-lead ECGs for home environments and multiple ECG comparison scenarios commonly encountered in clinical practice.
Furthermore, we propose the anyECG-chat model, which supports dynamic-length ECG inputs and multiple ECG inputs. We trained the model using a three-stage curriculum training recipe with the anyECG dataset.
A comprehensive evaluation was conducted, demonstrating that anyECG-chat is capable of supporting various practical application scenarios, including not only common report generation tasks but also abnormal waveform localization for long-duration reduced-lead ECGs in home environments and comprehensive comparative analysis of multiple ECGs. 
Our code and data are available at: https://github.com/CuCl-2/anyECG-chat
\end{abstract}

\begin{table}[t]
\centering
\small
\caption{anyECG dataset examples.}
\begin{tabular}{@{}p{1.0\linewidth}@{}}
\toprule
\multicolumn{1}{c}{\textbf{ReportGen}} \\
\midrule
\textbf{ECG (10s, 12leads)}: \\
\includegraphics[width=0.7\linewidth]{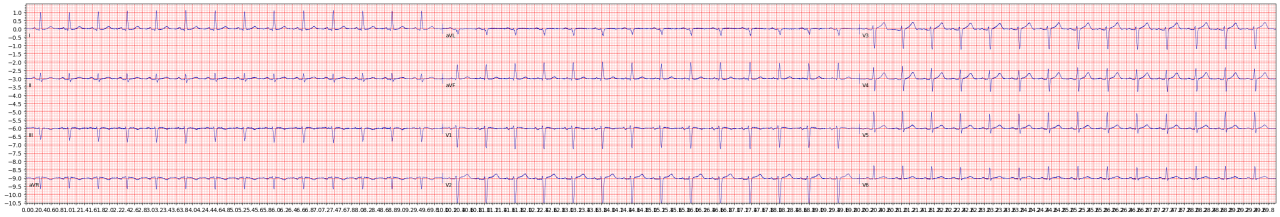} \\
\textbf{Q}: Please provide the report for the following ECG. \\
\textbf{A}: Report: Sinus rhythm, Poor R wave progression - probable normal variant, Borderline ECG \\
\midrule
\multicolumn{1}{c}{\textbf{Localization}} \\
\midrule
\textbf{ECG (15s, 2leads)}: \\
\includegraphics[width=0.7\linewidth]{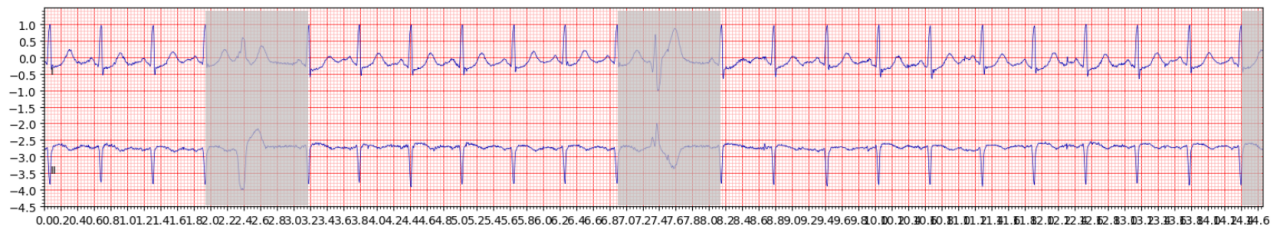} \\
\textbf{Q}: Examine this ECG and point out where the Premature ventricular contraction is located. \\
\textbf{A}: Duration: 1.9s-3.1s, 6.8s-8.1s, 14.3s-15.0s \\ 
\midrule
\multicolumn{1}{c}{\textbf{MultiECG}} \\
\midrule
\textbf{Three ECGs (10s, 12leads)}: \\
\includegraphics[width=0.7\linewidth]{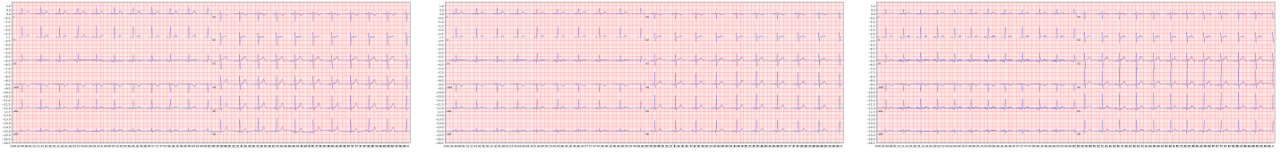} \\
\textbf{Q}: How do the ECGs collected 0 days ago, 1323 days ago, and 1924 days ago compare in terms of rhythm and overall assessment? \\
\textbf{A}: The ECGs collected at these times show sinus rhythm as the primary rhythm, with one instance of sinus bradycardia, and all are assessed as normal or normal except for rate, indicating a stable cardiac condition over time. \\
\bottomrule
\end{tabular}
\label{tab:dataset_examples}
\end{table}

\section{Introduction}
Electrocardiograms (ECGs) play a pivotal role in clinical practice, serving as a cornerstone for diagnosing and monitoring various cardiac conditions \cite{sahoo2020machine,rath2021heart,ayano2022interpretable}. Traditional methods for ECG analysis often rely on task-specific models, which are typically constrained to single tasks such as arrhythmia detection or report generation \cite{wang2023adversarial,eldele2021time,zhang2022maefe,hu2023spatiotemporal,na2024guiding}. These approaches lack the flexibility to handle diverse tasks and do not support interactive human-machine communication. The emergence of multimodal large language models (MLLMs) \cite{liu2023llava,wang2024qwen2,alayrac2022flamingo,chen2024far} has introduced a transformative paradigm for ECG analysis. By leveraging their powerful natural language reasoning capabilities and multimodal perception abilities, MLLMs can perform a wide range of tasks while enabling natural language-based human interaction. Consequently, ECG-focused MLLMs have garnered significant attention as a promising research direction.

Previous research on ECG MLLMs has explored various approaches. A straightforward method involves using an external classifier or feature extractor to convert ECG signals into a series of textual labels, which are then passed to the LLM \cite{liu2023biosignal,oh2023ecg,yu2023zero}. However, this approach often results in significant information loss. Another line of work focuses on native ECG MLLMs, where ECG representations are directly fed into the LLM. For instance, MEIT \cite{wan2024meit} employs a projection mechanism to align ECG embeddings with the semantic space of the LLM, enabling the generation of ECG reports. Similarly, PULSE \cite{liu2024teach} addresses real-world scenarios involving ECG images by constructing an ECG image-based MLLM that supports tasks such as report generation, waveform classification, and rhythm analysis. Building on these advancements, \cite{zhao2024ecg} introduces multi-turn dialogue capabilities, enabling iterative interactions. Additionally, it leverages retrieval-augmented generation (RAG) \cite{lewis2020retrieval} to enrich the LLM with ECG-specific knowledge.

Despite the advancements in ECG-MLLMs, their current applications are predominantly limited to single-task scenarios such as report generation or label classification \cite{wan2024meit,liu2024teach,zhao2024ecg,li2024biased}. In essence, ECG reports are composed of a series of labels related to rhythm, morphology, and diagnosis, making report generation and label classification fundamentally the same task \cite{gowmimic,wagner2020ptb}. However, the core objective of MLLMs is to address diverse, multi-task challenges rather than being limited to a single task \cite{wang2024qwen2}. Consequently, existing ECG-MLLMs fail to fully harness the potential of MLLMs. Moreover, these models are typically restricted to processing single, 12-lead, 10-second ECG inputs \cite{wan2024meit,liu2024teach,zhao2024ecg}, which are inadequate for modern use cases. For instance, they cannot effectively handle the long-duration, reduced-lead ECGs commonly generated in home environments \cite{gu2024effectiveness} or the multi-ECG comparison scenarios frequently encountered in clinical practice. To bridge this gap, there is a pressing need for a more versatile ECG-MLLM capable of supporting a broader range of tasks, particularly fine-grained localization tasks, and accommodating more flexible ECG inputs, including long-duration ECGs, reduced-lead ECGs, and multiple ECGs.

However, existing ECG question-answering datasets \cite{oh2023ecg,wan2024meit,liu2024teach} are often overly simplistic and fail to meet the requirements for diverse tasks and flexible input scenarios. To address these limitations, we developed a novel dataset named anyECG, which comprises three subsets: ReportGen, Localization, and MultiECG. These subsets encompass a wide range of tasks, including report generation, abnormal waveform localization, and open-ended question answering. Additionally, we introduced long-duration ECGs, reduced-lead ECGs, and multi-ECG inputs to better align with modern clinical and home-monitoring scenarios. 

To support these diverse tasks and flexible input formats, we propose the anyECG-chat Model which uses dynamic ECG input mechanism to support dynamic-length ECG inputs and multiple ECG inputs seamlessly. We employed a three-stage curriculum learning \cite{gong2023listen,wang2024qwen2} approach to train the model, enabling it to evolve from coarse perception to fine-grained understanding, and ultimately to instruction-following and multi-ECG comparison tasks.

We evaluated our model on three tasks. In the ReportGen task, out-of-domain testing on six unseen ECG datasets showed superior generalization compared to existing ECG-MLLMs. For the Localization task, using a reserved test set, our model outperformed traditional segmentation models and other ECG-MLLMs by enabling fine-grained, second-level abnormality localization and handling dynamic-length ECG inputs. It also showed strong zero-shot performance in unseen single-lead scenarios. In the MultiECG task, our model consistently led on the MIMIC Multi-ECG QA and ECG-QA datasets. Furthermore, our model demonstrated robust multi-turn dialogue capabilities.

Our contributions can be summarized as follows:
\begin{itemize}
  \item We introduce the anyECG dataset, which moves beyond traditional ECG report generation to fine-grained waveform localization and open-ended question answering. It also accommodates a wider variety of ECG input formats, including multi-ECG comparisons and long-duration, reduced-lead recordings.
  \item We proposed the anyECG-chat architecture, which is specifically designed to handle dynamic ECG inputs, enabling it to address the diverse scenarios presented by the anyECG dataset.
  \item We employed a three-stage curriculum learning approach, consisting of pre-training, fine-grained pre-training, and instruction tuning. The resulting anyECG-chat model demonstrates strong performance across various tasks, including report generation, waveform localization, and multi-ECG comparison.
\end{itemize}

\begin{figure*}[t]
  \centering
  \includegraphics[width=0.85\linewidth]{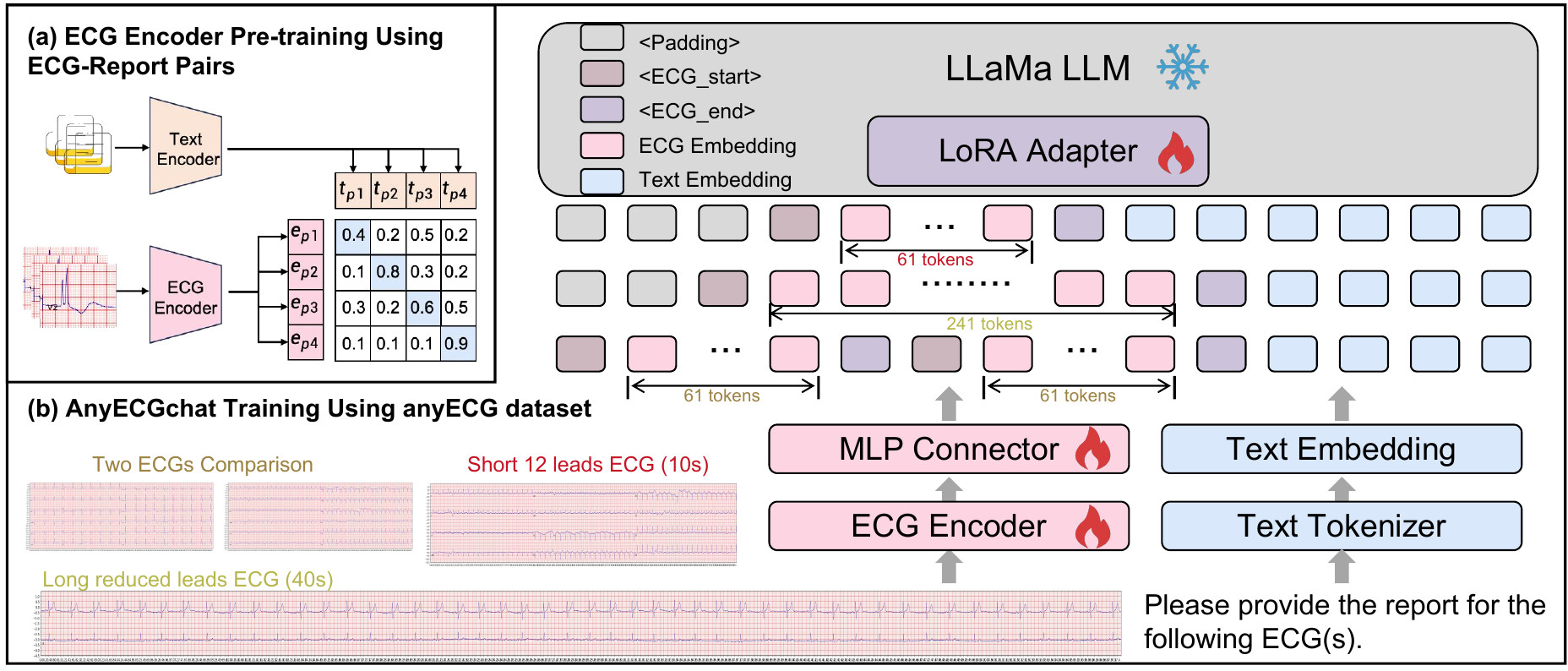}
  \caption{The overview of anyECG-chat architecture.}
  \label{fig:model_architecture}
\end{figure*}

\section{Related Work}
\textbf{ECG Understanding}: In recent years, the paradigm of ECG understanding has gradually shifted from traditional supervised learning \cite{ribeiro2020automatic} to self-supervised learning \cite{chen2020simple,grill2020bootstrap,chen2021exploring}, which leverages large amounts of unlabeled data for pretraining. Self-supervised ECG learning can be broadly categorized into contrastive self-supervised learning \cite{wang2023adversarial,eldele2021time} and generative self-supervised learning \cite{zhang2022maefe,hu2023spatiotemporal,na2024guiding}. Both approaches, however, require fine-tuning on downstream task data and are not inherently suited for zero-shot scenarios. Inspired by CLIP \cite{radford2021learning}, several multimodal contrastive learning methods for ECG-report pairs \cite{li2024frozen,liu2024zero,yu2024ecg,li2025fine} have emerged. However, these models lack a decoder and are therefore limited to discriminative tasks. They are ill-suited for diverse generative applications and cannot accommodate multiple tasks within a single model. In contrast, this paper introduces anyECG-chat, a generative MLLM capable of performing a wide range of tasks guided by textual instructions. This approach unlocks the potential for diverse and flexible applications in ECG understanding.

\textbf{ECG-MLLMs} Inspired by advancements in large vision-language models \cite{liu2023llava,wang2024qwen2,alayrac2022flamingo,chen2024far}, ECG-MLLMs have emerged as a promising direction for ECG analysis. A straightforward approach involves using external classifiers or feature extractors to convert ECG signals into a series of textual labels, which are then fed into the LLM \cite{liu2023biosignal,oh2023ecg,yu2023zero}. However, this method often results in significant information loss. An alternative approach like MEIT \cite{wan2024meit} focuses on building native ECG MLLMs, where ECG embeddings are directly input into the LLM. PULSE \cite{liu2024teach} addresses real-world scenarios involving ECG images by constructing an ECG image-based MLLM capable of supporting tasks such as report generation, waveform classification, and rhythm analysis. Building on these advancements, \cite{zhao2024ecg} introduces multi-turn dialogue capabilities, allowing for iterative interactions. Additionally, it incorporates RAG \cite{lewis2020retrieval} to enhance the LLM with ECG-specific knowledge.

Despite these successes, existing ECG-MLLMs are often limited to report generation applications or label classification \cite{wan2024meit,liu2024teach,zhao2024ecg} and are typically restricted to processing single, 12-lead, 10-second ECG inputs. This limitation makes them inadequate for modern scenarios. This paper aims to develop a more versatile ECG-MLLM that supports a broader range of tasks and accommodates more flexible ECG inputs.

\section{AnyECG Dataset}
The existing ECG-QA datasets \cite{oh2023ecg,wan2024meit,liu2024teach} are relatively monotonous and fail to meet the requirements for supporting multi-task and flexible ECG inputs. To address this limitation, we constructed a novel dataset named anyECG. In terms of tasks, prior ECG-QA datasets are often restricted to report generation or label classification, which significantly underestimates the potential of MLLMs. Therefore, we introduced a broader range of tasks, including not only report generation but also more complex tasks requiring fine-grained perception, such as waveform localization, as well as diverse open-ended question answering facilitated by LLMs. Regarding ECG signals, previous datasets typically utilize single, short-duration (10s), 12-lead ECG. This setup is inadequate for modern scenarios, such as the large volume of long-duration, reduced-lead ECGs generated in home environments, and the multi-ECG comparison scenarios commonly encountered in clinical practice. To this end, we incorporated long-duration ECGs, reduced-lead ECGs, and multi-ECG inputs into our dataset. Specifically, anyECG consists of three components: anyECG-ReportGen, anyECG-Localization, and anyECG-MultiECG. We provide an example from each component in the Table~\ref{tab:dataset_examples} and summarize the dataset statistics in Table \ref{tab:dataset_statistic}. Notably, we standardized the sampling frequency of all ECG datasets to 100 Hz and normalized the ECG signals to a range of -1 to 1.

\begin{table}[t]
  \centering
  \caption{Overview of the anyECG Dataset. ESTD: European ST-T Database, MIT-ST: MIT-BIH ST Change Database, MIT-Arr: MIT-BIT Arrhythmia Database.}
  \resizebox{\linewidth}{!}{
  \begin{tabular}{@{}llcccc@{}}
  \toprule
  \textbf{Dataset} & \textbf{Source} & \textbf{Duration} & \textbf{Leads} & \textbf{ECGs per QA} & \textbf{QA Pairs} \\ \midrule
  \multicolumn{6}{c}{\textbf{ReportGen}} \\ \midrule
  MIMIC-ECG ReportGen & MIMIC & 10s & 12 & 1 & 773,268 \\ \midrule
  \multicolumn{6}{c}{\textbf{Localization}} \\ \midrule
  ESTD Loc & ESTD & 10s & 2 & 1 & 39,110 \\ 
  ESTD Loc Long & ESTD & 10--60s & 2 & 1 & 19,555 \\ 
  MIT-ST Loc & MIT-ST & 10s & 2 & 1 & 6,500 \\ 
  MIT-ST Loc Long & MIT-ST & 10--60s & 2 & 1 & 3,250 \\ 
  MIT-Arr Loc & MIT-Arr & 10s & 2 & 1 & 54,440 \\ 
  MIT-Arr Loc Long & MIT-Arr & 10--60s & 2 & 1 & 27,220 \\ \midrule
  \multicolumn{6}{c}{\textbf{MultiECG}} \\ \midrule
  MIMIC Multi-ECG QA & MIMIC & 10s & 12 & 2--6 & 135,094 \\ 
  ECG-QA(10\%) & PTB-XL & 10s & 12 & 1--2 & 33,220 \\ \bottomrule
  \end{tabular}
  }
  \label{tab:dataset_statistic}
\end{table}

Our anyECG dataset is constructed by reorganizing existing datasets or with the assistance of LLM-generated data. Specifically, anyECG-ReportGen is constructed by reorganizing the MIMIC-ECG \cite{gowmimic} dataset, anyECG-Localization is built by reorganizing three long-duration, 2-lead ECG datasets: the European ST-T Database \cite{taddei1992european}, the MIT-BIH ST Change Database \cite{albrecht1983st}, and the MIT-BIT Arrhythmia Database \cite{moody2001impact}. anyECG-MultiECG is constructed based on the MIMIC-ECG \cite{gowmimic} and PTB-XL \cite{wagner2020ptb} datasets. The detailed construction process is described in Appendix AnyECG Dataset Construction.

\section{AnyECG-chat Architecture}
The architecture of our model is illustrated in Figure \ref{fig:model_architecture}. It consists of an ECG encoder, a large language model (LLM), a modality alignment module, and LoRA adapters. Previous ECG MLLMs were often limited to single 12-lead, short-duration (10s) ECG inputs. To enable our model to handle the diverse scenarios and flexible ECG inputs in the anyECG dataset, we introduced a Dynamic ECG Input mechanism. We will elaborate on each component and the Dynamic ECG Input mechanism in detail below.

\subsection{ECG encoder} 
The performance of multimodal large language models (MLLMs) in question answering (QA) tasks heavily relies on the perceptual capabilities of the ECG encoder. Instead of training the ECG encoder from scratch, we opted to pre-train it using contrastive learning \cite{li2024frozen} on the MIMIC-ECG \cite{gowmimic} dataset, which contains 800,000 ECGs and their corresponding reports.

We employed a ViT-base \cite{dosovitskiy2020image} architecture as the ECG encoder. However, since ViT-base is originally designed for image data, we redefined the patching mechanism to accommodate the temporal and multi-lead nature of ECG signals; and introduced lead embeddings \cite{na2024guiding}, adjusted positional embeddings to capture the spatiotemporal structure of ECG data.

Specifically, let an ECG signal be represented as \( X \in \mathbb{R}^{L \times T} \), where \( L \) is the number of leads and \( T \) is the signal length. First, we standardized the sampling frequency to 100 Hz and normalized each lead to the range \([-1, 1]\) to mitigate measurement biases from different devices and enhance generalization. To adapt the patching mechanism, we applied spatio-temporal patchifying with a patch size of \( (1, 200) \). For example, given a preprocessed ECG from MIMIC-ECG \( X \in \mathbb{R}^{12 \times 1000} \), each lead is divided into 5 patches, resulting in a total of 60 patches across all leads. Additionally, we introduced a [CLS] token to capture global features.

Traditional ViT models rely solely on positional embeddings, which are insufficient for capturing inter-lead relationships in ECG data. To address this limitation, we introduced lead embeddings, denoted as \( E_{\text{lead}} \), to encode the spatial relationships between leads. Patches from the same lead share the same lead embedding, while patches from different leads at the same time share the same positional embedding. The final input embedding for each patch is computed as:

\[
E = E_{\text{signal}} + E_{\text{pos}} + E_{\text{lead}},
\]
where \( E_{\text{pos}} \) represents the positional embedding, and \( E_{\text{signal}} \) is the patch embedding derived from the ECG signal.

\begin{table*}[t]
  \centering
  \small
  \caption{Overview of the Training Recipe}
  \resizebox{0.8\linewidth}{!}{
  \begin{tabular}{@{}ccccccc@{}}
  \toprule
  \textbf{Stage} & \textbf{Trained Params} & \textbf{Training Task} & \textbf{Samples} & \textbf{LR} & \textbf{Batch Size} & \textbf{Epochs} \\ \midrule
  1 & Connector + ECG encoder & ReportGen & 773,268 & $1 \times 10^{-4}$ & 256 & 2 \\
  2 & Connector + ECG encoder + LoRA & ReportGen + Localization & 923,343 & $1 \times 10^{-4}$ & 64 & 2 \\
  3 & Connector + LoRA & ReportGen + Localization + MultiECG & 1,091,657 & $1 \times 10^{-4}$ & 64 & 1 \\ \bottomrule
  \end{tabular}
  }
  \label{tab:training_recipe}
  \end{table*}

\subsection{Large Language Model.}
In this paper, we utilize the Meta-Llama-3-8B-Instruct \cite{grattafiori2024llama} as our LLM. To prevent overfitting and catastrophic forgetting, which could significantly degrade the model's ability to respond to general queries, we opted for Low-Rank Adaptation (LoRA) \cite{hu2021lora} instead of full parameter fine-tuning. In anyECG-chat, we inject LoRA adapters (rank=8 and $\alpha=16$) to the projection layers for query and key in all self-attention layers of the LLaMA model.

\subsection{Modality Connector}
Various modality connectors have been explored in prior research on vision-language models (VLMs), including cross-attention mechanisms \cite{alayrac2022flamingo}, Q-formers \cite{li2023blip}, and simple linear projections \cite{liu2023llava}. In this work, to balance effectiveness and efficiency, we adopt a two-layer MLP with GELU activation as the modality connector, inspired by LLaVA 1.5 \cite{liu2023improvedllava}.

\subsection{Dynamic ECG Input}
The Dynamic ECG Input mechanism is designed to empower anyECG-chat with the ability to handle diverse scenarios and flexible ECG inputs, including varying-length ECGs, reduced-lead ECGs, and multi-ECG inputs. To achieve this, two key challenges must be addressed: (1) embedding dynamic-length and reduced-lead ECGs effectively, and (2) ensuring that multiple ECG embeddings can be input into the LLM while maintaining clear distinctions between different ECGs.

For the first challenge, since our ECG encoder is pre-trained on the MIMIC-ECG dataset using 10-second, 100 Hz, 12-lead ECGs, we adopt the following strategies: For ECGs shorter than 10 seconds, zero-padding is applied to match the required length. For ECGs longer than 10 seconds, they are first padded to the nearest multiple of 10 seconds and then segmented into 10-second clips. These clips are individually processed by the ECG encoder, and the resulting embeddings are concatenated to get the final ECG embedding sequences. The \texttt{[CLS]} tokens from each segment are averaged to produce the final \texttt{[CLS]} embedding for the long-duration ECG. For reduced-lead ECGs, missing leads are similarly zero-padded to ensure compatibility with the encoder. As mentioned above, since our ECG encoder incorporates lead embeddings, it can capture the relationships between leads even for missing leads.

To address the second challenge, and to ensure the LLM can distinguish between multiple ECG inputs without conflating them into a single long-duration ECG, we introduce special tokens \texttt{<ECG\_start>} and \texttt{<ECG\_end>}. These tokens are added before and after each ECG embedding, enabling the LLM to clearly identify and differentiate between individual ECG inputs.

\section{Training Recipe}
We designed a three-stage curriculum learning approach tailored to the varying complexity of tasks in the anyECG dataset. Inspired by \cite{gong2023listen,wang2024qwen2}, this approach comprises pretraining, fine-grained pretraining, and open-ended instruction tuning. Notably, the ECG encoder was pre-trained on the MIMIC-ECG dataset using contrastive learning prior to these three stages.

In Stage 1, the model was trained on the anyECG-ReportGen dataset with a frozen LLM; only the ECG encoder and Connector were updated to align ECG and LLM embeddings. Stage 2 added the more demanding anyECG-Localization dataset, requiring fine-grained waveform localization. Here, we jointly trained the ECG encoder, Connector, and fine-tuned the LLM using LoRA to enhance localization performance. In Stage 3, we introduced open-ended QA tasks using the full anyECG dataset, freezing the ECG encoder. This phase emphasized instruction-following and incorporated multi-ECG inputs for comparative reasoning. Notably, the dataset for each stage includes the dataset from the previous stage, preventing the model from forgetting. This progressive three-stage training strategy allowed the model to evolve from coarse perception to fine-grained understanding, and finally to instruction-following and multi-ECG comparison tasks. By gradually increasing task complexity, the approach mitigates the risk of the model relying excessively on textual reasoning, which could lead to hallucinations, especially when its ECG perception capabilities are underdeveloped. The training recipe and detailed hypyerparameters are summarized in Table \ref{tab:training_recipe}.

\section{Experiments}
We evaluated the performance of anyECG-chat across three tasks. For Report Generation task, since MIMIC-ECG dataset was used for contrastive pretraining and Stage 1 training, we performed out-of-domain testing on six unseen ECG datasets to ensure fairness. For Localization, we used the reserved test set from anyECG-Localization to evaluate the model's performance and further assessed its zero-shot capability in single-lead scenarios. For Multi-ECG, we evaluated the model on the reserved test set of the MIMIC Multi-ECG QA dataset and the ECG-QA dataset, which contains multi-turn question-answering tasks. We also conducted a qualitative analysis of multi-turn instruction-following capabilities. An overview of the evaluation datasets is provided in Appendix Evaluation Dataset Overview.

\subsection{Report Generation}
As discussed above, we used six OOD ECG classification datasets to evaluate the generalization capability of anyECG-chat. Notably, ECG reports are essentially composed of labels, and using classification metrics to evaluate the model provides a more accurate measure of its understanding of ECGs compared to traditional text similarity metrics like BLEU or ROUGE. An example is that predition `sinus tachycardia' and ground truth `sinus bradycardia' would yield a classification score of 0, while semantic similarity metrics might still assign a non-zero score. 

To compare anyECG-chat with existing models, we prompted the anyECG-chat with the query, ``Please provide the report for the following ECG." The reports generated by anyECG-chat and the dataset label names were then encoded using a text encoder (BioBERT \cite{deka2022evidence}). Finally, the cosine similarity between the text embeddings of anyECG-chat's output and each label was computed to derive the prediction scores.

We compared anyECG-chat against several supervised methods \cite{wang2023adversarial,na2024guiding}, discriminative zero-shot methods \cite{liu2024zero}, and other generative zero-shot methods \cite{li2023llava,liu2024teach,wan2024meit} using AUC as the evaluation metric. The results, presented in Table~\ref{tab:results_classification}, demonstrate that anyECG-chat achieved the best performance among generative zero-shot methods like PULSE \cite{liu2024teach} and MEIT \cite{wan2024meit}. Although our method does not outperform discriminative zero-shot methods, such a comparison is actually unfair because MERL \cite{liu2024zero} uses the labels of each dataset as prior knowledge, whereas our model does not require any label information. Our method directly generates labels without relying on prior knowledge. Despite this unfair comparison, our model still achieves comparable performance on PTBXL-Rhythm, PTBXL-Sub, and CSN. We also present the results using semantic similarity metrics as a reference in Table~\ref{tab:report_gen_results}. The reported metrics are averaged across the six datasets.

\subsection{Localization Task}
\subsubsection{Results}
For the localization task, we used the reserved test set from anyECG-Localization to evaluate the model's performance using the Intersection over Union (IoU) metric. We also compared its performance against other supervised methods \cite{moskalenko2020deep} and existing ECG-MLLMs. As expected, other ECG-MLLMs lacked the fine-grained temporal perception required for second-level localization. When asked to identify the location of abnormal waveforms, they could only provide lead-level answers (see Appendix Case Study for detail).  Detailed results are presented in Figure~\ref{fig:results_location}.

For short-duration ECGs, the results demonstrate that although Unet \cite{moskalenko2020deep} is a dedicated model for segmentation tasks, anyECG-chat significantly outperformed Unet on the European ST-T and MIT-BIH ST Change datasets and achieved comparable performance on the MIT-BIH Arrhythmia dataset. Interestingly, we observed that Unet exhibited consistent performance across different datasets, whereas anyECG-chat showed varying performance. This discrepancy may be attributed to the diverse training data used for anyECG-chat, beyond the anyECG-localization dataset, which likely enhanced its ability to perceive different types of abnormalities across datasets. For long-duration ECGs, Unet was unable to handle dynamic-length ECGs due to architectural limitations, whereas anyECG-chat successfully processed these inputs, further showcasing its flexibility and robustness.

\subsubsection{Zero-shot Single Lead Localization}
\begin{figure}[t]
  \centering
  \includegraphics[width=\linewidth]{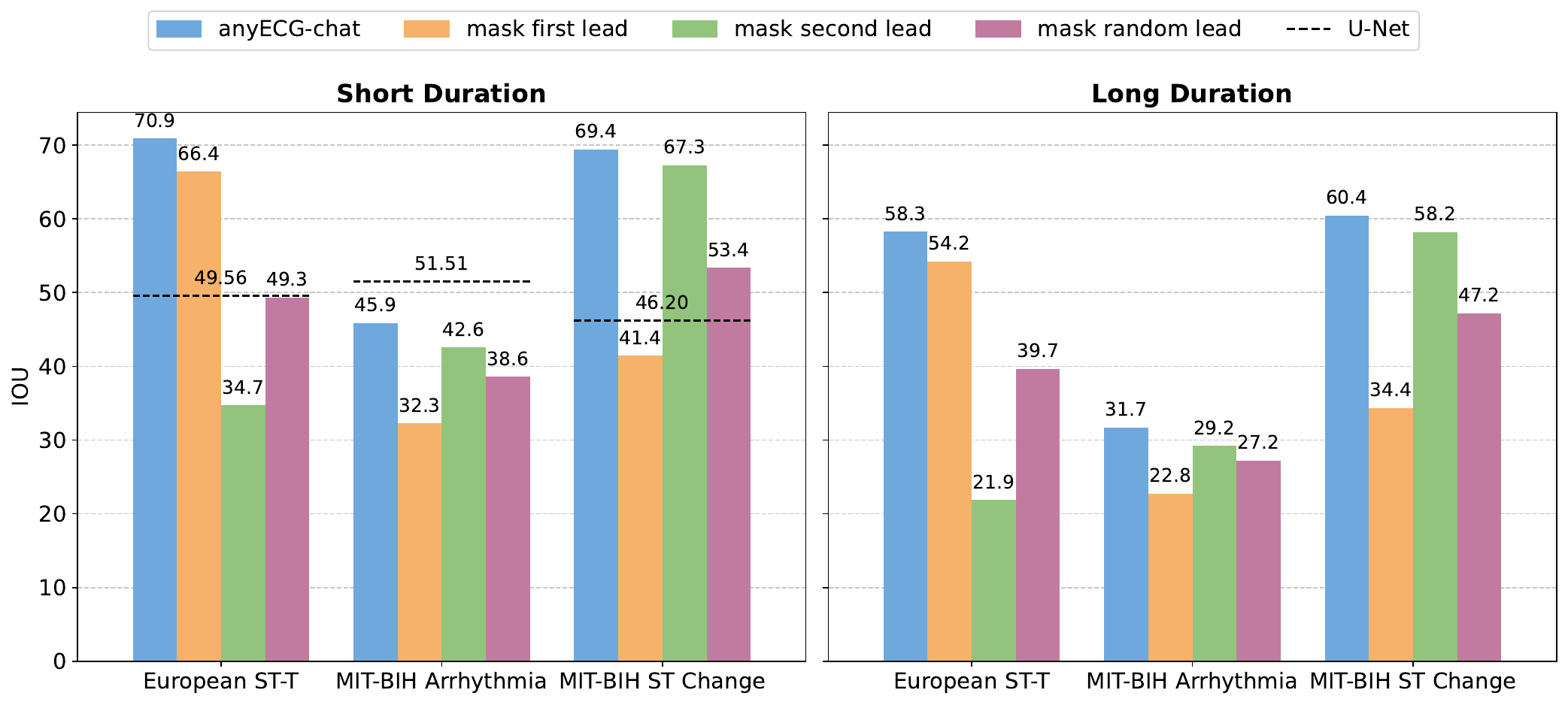}
  \caption{Results of Localization and Zero-Shot Single Lead ECG Localization. \textit{Since LLaVa-Med, MEIT and PULSE failed to provide second-level localization, scoring 0, they are omitted from the figure}.}
  \label{fig:results_location}
\end{figure}

Though anyECG-localization dataset only includes 2-lead ECGs, we also evaluated the model's zero-shot capability in single-lead scenarios. Three single-lead cases were tested: masking the first lead, masking the second lead, and masking a random lead, with the masked lead values set to zero. The results, shown in Figure~\ref{fig:results_location}, indicate that anyECG-chat achieves comparable performance in both short-duration and long-duration ECGs when the first lead is masked in the European ST-T dataset and when the second lead is masked in the MIT-BIH ST Change and MIT-BIH Arrhythmia datasets. This demonstrates the model's zero-shot capability in single-lead scenarios. However, performance drops significantly when the other lead is masked, likely because the queried abnormal waveform features are present only in the masked lead.

\begin{table}[t]
  \centering
  \caption{Results of Classification.}
  \label{tab:results_classification}
  \resizebox{\linewidth}{!}{%
  \begin{tabular}{lcccccc}
      \toprule
      \textbf{macro-AUC} & \makecell{\textbf{PTBXL}\\\textbf{Super}} & \makecell{\textbf{PTBXL}\\\textbf{Sub}} & \makecell{\textbf{PTBXL}\\\textbf{Form}} & \makecell{\textbf{PTBXL}\\\textbf{Rhythm}} & \textbf{CPSC} & \textbf{CSN} \\
      \midrule
      \multicolumn{7}{l}{\textcolor{gray}{Supervised: \textit{dedicated model tailored for each dataset}}} \\
      ASTCL & 81.02 & 76.51 & 66.99 & 76.05 & 79.51 & 75.79 \\
      ST-MEM & 71.36 & 63.59 & 66.07 & 74.85 & 70.39 & 71.36 \\
      \midrule
      \multicolumn{7}{l}{\textcolor{gray}{Discriminative Zero-Shot: \textit{requires pre-defined labels}}} \\
      MERL & 74.20 & 75.70 & 65.90 & 78.50 & 82.80 & 74.40 \\
      \midrule
      \multicolumn{7}{l}{\textcolor{gray}{Generative Zero-Shot: \textit{directly outputs labels}}} \\
      LLaVa-Med & 51.21 & 58.33 & \textbf{69.12} & 75.77 & 56.07 & 60.54 \\
      MEIT & 62.34 & 57.91 & 61.12 & 70.45 & 62.38 & 62.73 \\
      PULSE & 66.61 & 61.32 & 63.82 & 73.91 & 66.15 & 64.18 \\
      anyECG-chat & \textbf{68.95} & \textbf{73.10} & 64.55 & \textbf{77.60} & \textbf{71.05} & \textbf{71.29} \\
      \bottomrule
  \end{tabular}
  }
\end{table}

\begin{table}[t]
\centering
\caption{Results of Report Generation}
\setlength{\tabcolsep}{3pt}  
\resizebox{\linewidth}{!}{%
\begin{tabular}{lccccccc}
\toprule
\small\textbf{Model} & \small\textbf{BLEU-1} & \small\textbf{BLEU-2} & \small\textbf{BLEU-3} & \small\textbf{BLEU-4} & \small\textbf{ROUGE-1} & \small\textbf{ROUGE-2} & \small\textbf{ROUGE-L} \\
\midrule
LLaVa-Med & 0.47 & 0.44 & 0.40 & 0.38 & 0.62 & 0.57 & 0.52 \\
MEIT & 0.47 & 0.43 & 0.40 & 0.37 & 0.64 & 0.60 & 0.54 \\
PULSE & 0.50 & 0.46 & 0.43 & 0.40 & 0.68 & 0.64 & 0.58 \\
anyECG-chat & \textbf{0.53} & \textbf{0.51} & \textbf{0.47} & \textbf{0.44} & \textbf{0.72} & \textbf{0.68} & \textbf{0.60} \\
\bottomrule
\end{tabular}%
}
\label{tab:report_gen_results}
\end{table}

\subsection{Multi-ECG Comparison}
For the multi-ECG comparison task, we evaluated our model using two datasets: MIMIC Multi-ECG QA and ECG-QA. The former includes scenarios involving comparisons of 2 to 6 ECGs, while the latter focuses solely on comparisons between 2 ECGs. As previously mentioned, since the answers in ECG-QA are relatively concise, we limited the training data to 10\% of the original dataset to prevent the model from overfitting to short responses.

\subsubsection{MIMIC Multi-ECG QA}
Since the MIMIC Multi-ECG QA dataset is constructed using Llama-3.3-70B-Instruct \cite{touvron2023llama,grattafiori2024llama} for open-ended QA tasks, it lacks explicit metrics for direct evaluation. To address this, we employed QwQ-32B \cite{qwq32b,qwen2.5}, as the evaluation model. To ensure fairness, we did not use the answer generated by Llama-3.3-70B-Instruct as the gold standard for QwQ's evaluation. Instead, we provided QwQ with the questions and the corresponding reports for each ECG, allowing it to assess the quality of the outputs without bias. The evaluation scores ranged from 0 to 5. The detailed prompt is provided in Appendix Case Study.
Furthermore, to further validate the accuracy of LLM-based evaluation, we sampled 120 data points for human scoring, as detailed in Appendix Human Scoring, which demonstrates a strong correlation between human scores and QwQ's scores.

\begin{figure}[t]
\centering
\includegraphics[width=\linewidth]{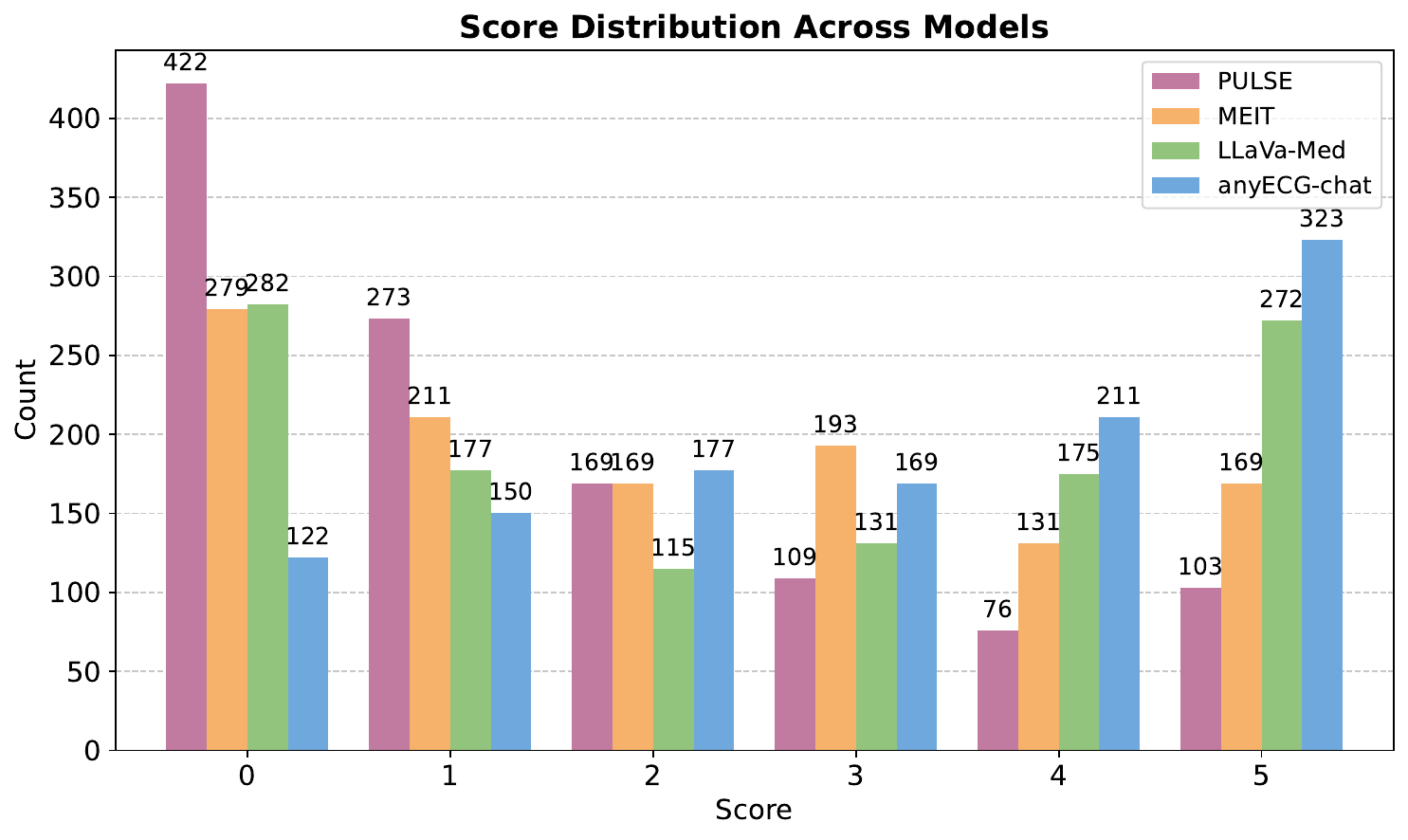}
\caption{The Score Distribution on MIMIC Multi-ECG QA.}
\label{fig:results_mimic}
\end{figure}

We compared the outputs of anyECG-chat and other ECG-MLLMs. It is worth noting that since LLaVa-Med, MEIT and PULSE were not trained to handle multi-ECG inputs, we adapted their usage to support multi-ECG comparison tasks while maintaining consistency with their training setup. Specifically, we first processed each ECG individually to generate its corresponding report. These reports were then concatenated, along with an image combining all the ECGs, and provided as input to ECG-MLLMs to answer multi-ECG comparison questions. The score distributions for these models are shown in Figure~\ref{fig:results_mimic}. Notably, anyECG-chat achieved significantly higher scores compared to the other two models. Additionally, we analyzed the average scores of each model across different numbers of ECG inputs, as well as the number of times each model achieved the highest score among the three models. The results, as shown in Table~\ref{tab:results_mimic_detail}, indicate that anyECG-chat maintains notable robustness as the number of input ECGs increases. Furthermore, anyECG-chat secured the highest score in 816 out of 1,152 questions, demonstrating a substantial performance advantage over the other two models.

\subsubsection{ECG-QA}
For the ECG-QA dataset, we compared anyECG-chat with several discriminative models \cite{chen2022multi,moon2022multi} and other ECG-MLLMs \cite{liu2024teach,wan2024meit}. As mentioned earlier, the answers in ECG-QA are relatively concise, often limited to a few short phrases. Consequently, discriminative methods model the QA task as a multi-label classification problem, which requires predefined possible labels as prior knowledge. In contrast, ECG-MLLMs, as generative methods, directly produce answers without relying on predefined labels. We used exact match accuracy as the evaluation metric, and the results are presented in Table~\ref{tab:results_ecgqa}. Although anyECG-chat does not outperform discriminative models that leverage predefined labels, it achieves the best performance among generative ECG-MLLMs, even when trained on only 10\% of the training data. Notably, it excels in CI-Verify and CC-Verify tasks, achieving accuracies of 70.1\% and 67.9\%, respectively, demonstrating its strong capability in multi-ECG comparison tasks.

\begin{table}[t]
\centering
\caption{Average Scores for MIMIC Multi-ECG QA Across Different Numbers of ECG Inputs.}
\resizebox{\linewidth}{!}{%
\begin{tabular}{@{}cccccccc@{}}
\toprule
\textbf{\# ECGs} & \textbf{2} & \textbf{3} & \textbf{4} & \textbf{5} & \textbf{6} & \textbf{All} & \textbf{Highest (\#)}\\ \midrule
PULSE & 1.66 & 1.40 & 1.13 & 1.46 & 1.84 & 1.53 & 264/1152\\
MEIT & 2.10 & 1.85 & 1.75 & 1.80 & 2.00 & 1.90 & 415/1152 \\
LLaVa-Med & 2.75 & 2.19 & 2.34 & 2.09 & 2.24 & 2.48 & 553/1152 \\
anyECG-chat & 3.30 & 2.81 & 2.59 & 2.63 & 2.98 & 3.02 & 816/1152\\\bottomrule
\end{tabular}
}
\label{tab:results_mimic_detail}
\end{table}

\begin{table}[t]
  \centering
  \caption{Performance Comparison on ECG-QA.}
  \setlength{\tabcolsep}{3pt}  
  \resizebox{\linewidth}{!}{%
  \begin{tabular}{@{}lccccccc@{}}
    \toprule
    \textbf{EM Acc.} & \makecell{\textbf{S}\\\textbf{Verify}} & \makecell{\textbf{S}\\\textbf{Choose}} & \makecell{\textbf{S}\\\textbf{Query}} & \makecell{\textbf{CC}\\\textbf{Verify}} & \makecell{\textbf{CC}\\\textbf{Query}} & \makecell{\textbf{CI}\\\textbf{Verify}} & \makecell{\textbf{CI}\\\textbf{Query}} \\ 
    \midrule
    \multicolumn{8}{l}{\textcolor{gray}{Discriminative Model: \textit{requires possible labels}}} \\
    M\textsuperscript{3}AE & 74.6 & 57.1 & 41.0 & 75.5 & 20.1 & 75.3 & 4.2 \\ 
    MedViLL & 73.9 & 54.1 & 40.4 & 74.3 & 22.0 & 77.5 & 3.5 \\ 
    Fusion Transformer & 72.1 & 46.4 & 37.4 & 71.9 & 18.4 & 68.1 & 2.2 \\ 
    \multicolumn{8}{l}{\textcolor{gray}{Generative ECG-MLLM: \textit{directly outputs answers}}} \\
    LLaVa-Med(0\%) & 34.7 & 0 & 0 & 11.9 & 0 & 36.8 & 0 \\
    MEIT(0\%) & 42.2 & 2.5 & 0.6 & 28.3 & 0.4 & 40.5 & 0 \\
    PULSE(100\%) & 64.6 & 56.1 & 2.4 & 52.9 & 3.9 & 57.1 & 0 \\
    anyECG-chat(10\%) & 69.6 & 50.1 & 20.1 & 68.0 & 8.6 & 72.1 & 1.2 \\
    \bottomrule
  \end{tabular}
  }
  \begin{tablenotes}
    \small
    \item \tiny S: Single, CC: Comparison-Consecutive, CI: Comparison-Irrelevant.
  \end{tablenotes}
  \label{tab:results_ecgqa}
\end{table}

\begin{table}[t]
\centering
\caption{Ablation Study Results.}
\setlength{\tabcolsep}{3pt}  
\resizebox{\linewidth}{!}{%
\begin{tabular}{@{}lccc@{}}
\toprule
\textbf{Configuration} & \textbf{Classification} & \textbf{Localization} & \textbf{MultiECG} \\ \midrule
Default & 71.09 & 56.10 & 3.02 \\
w/o Contrastive Pre-training & 68.24 & 53.69 & 2.82 \\
w/o Lead Embedding & 70.10 & 55.53 & 2.93 \\
w/o Dynamic Input Mechanism & 70.55 & 52.66 & 2.74 \\
\midrule
Full Parameter Tuning & 71.02 & 56.15 & 3.04 \\
w/o Experience Replay & 66.32 & 53.88 & 3.04 \\
w/o Curriculum Training & 61.21 & 50.62 & 2.95 \\ \bottomrule
\end{tabular}%
}
\label{tab:ablation_study}
\end{table}

\subsection{Ablation Study and Analysis}
\subsubsection{Model Architecture.} We performed a series of ablation studies, the results are summarized in Table \ref{tab:ablation_study}. The reported metrics represent the average performance on the three dimensions. The ablation studies reveal several critical insights. First, initializing the ECG encoder with random weights instead of using multimodal contrastive pre-training significantly degrades performance across all three tasks, underscoring the importance of pre-trained representations in capturing rich ECG features. Second, removing lead embedding results in slight performance drops, particularly in tasks requiring complex spatial relationships, such as localization and MultiECG. Third, omitting the dynamic input mechanism restricts the model to processing fixed 10-second ECG inputs. While short ECG classification tasks remain largely unaffected, localization and MultiECG performance suffer due to the inability to handle long-duration or multi-segment ECGs.

\subsubsection{Training Strategy.} Fully fine-tuning all parameters instead of using LoRA substantially increases computational cost without noticeable performance improvements, highlighting the efficiency of LoRA-based adaptation. When experience replay was removed, where earlier data and tasks were excluded from subsequent training stages. This approach led to noticeable forgetting of previously learned information \cite{rolnick2019experience,scialom2022fine}. Additionally, when curriculum training was eliminated, the three-stage training pipeline was replaced with a single-stage training approach. This resulted in a substantial decline in performance across both Classification and Localization tasks. The drop can be attributed to the inherent complexity of the MultiECG task, which involves handling open-ended question answering. Directly starting training on such high-difficulty data caused the model to rely heavily on its language modeling capabilities, leading to increased hallucination.

\subsubsection{Multi-Turn QA.} Although anyECG dataset contains only single-turn QA, we hypothesize that anyECG-chat can handle multi-turn QA due to LoRA-based fine-tuning, which preserves the LLM’s pre-trained abilities. As illustrated in Appendix Multi-Turn QA, the model shows strong multi-turn instruction-following behavior, indicating its potential as a teaching aid for physicians despite the lack of quantitative evaluation.

\section{Conclusion}
In this paper, we introduced anyECG-chat, a MLLM designed for diverse ECG analysis tasks. By leveraging the novel anyECG dataset and a three-stage curriculum training strategy, anyECG-chat demonstrated strong performance across report generation, waveform localization, and multi-ECG comparison tasks. The proposed Dynamic ECG Input mechanism further enhanced the model's flexibility, enabling it to handle varying-length, reduced-lead, and multi-ECG inputs seamlessly. Experimental results showed that anyECG-chat outperformed existing ECG-MLLMs in multiple scenarios and exhibited robust zero-shot capabilities. 

\section{Acknowledgments}
This work was partially supported by the Technical Innovation key project of Zhejiang Province (2024C03023), and the National Nature Science Foundation of China under Grant No. 82272129 and 82572380.

\bibliography{aaai2026}

\clearpage
\section{AnyECG Dataset Construction}
\label{sec:appendix_dataset_construction}
The detailed construction process of the anyECG dataset is described below, including the preprocessing steps for each component and the generation of QA pairs.

\subsection{anyECG-ReportGen}
anyECG-ReportGen is a report generation QA dataset derived from the MIMIC-ECG dataset. MIMIC-ECG \cite{gowmimic} contains approximately 800,000 ECGs and corresponding reports collected from around 160,000 individuals. Each ECG consists of 12 leads, has a duration of 10 seconds, and is sampled at 500 Hz. To enhance data quality, we excluded samples with empty reports or reports containing fewer than three words, removed reports lacking meaningful information, and discarded ECGs with unexpected anomalies. After these preprocessing steps, a total of 773,268 ECGs remained. We organized the ECGs and their corresponding reports into a QA format suitable for training MLLMs. The questions were formulated as "Please provide the report for the following ECG" and its various paraphrased forms (see Table \ref{tab:template_report_gen}). The answers correspond to the respective reports. In total, 773,268 QA pairs were generated, all of which were used for training.

\begin{tcolorbox}[colback=gray!10, colframe=black, boxrule=0.5pt, arc=3pt, width=\linewidth, title=Questions for ECG Report Generation.]
  \begin{itemize}[leftmargin=1.5em]
  \footnotesize
    \item Please provide the report for the following ECG.
    \item Give me the report of this ECG.
    \item I need a report on the following ECG.
    \item Could you send me the ECG report?
    \item Provide me with the report of this ECG.
    \item Please generate a report for the ECG below.
    \item I’d like to receive the report for this ECG.
    \item Can you share the report of the following ECG?
    \item Give me a detailed report on this ECG.
    \item May I have the official report for the ECG provided?
  \end{itemize}
\end{tcolorbox}
\captionsetup{type=table}
\captionof{table}{Questions for ECG Report Generation.}
\label{tab:template_report_gen}

\subsection{anyECG-Localization}
anyECG-Localization is a waveform localization dataset derived from three long-duration, 2-lead ECG datasets collected in home settings: the European ST-T Database \cite{taddei1992european}, the MIT-BIH ST Change Database \cite{albrecht1983st}, and the MIT-BIT Arrhythmia Database \cite{moody2001impact}. These datasets are meticulously annotated by physicians to identify abnormal waveforms and rhythms, including features such as Left Bundle Branch Block (LBBB) beats, Right Bundle Branch Block (RBBB) beats, and Premature Ventricular Contractions (PVCs). Specifically, the European ST-T Database contains 90 ECG recordings, each lasting 120 minutes. The MIT-BIH ST Change Database includes 28 ECG recordings, each lasting between 20 and 70 minutes. The MIT-BIT Arrhythmia Database comprises 48 ECG recordings, each lasting 30 minutes. While all these datasets consist of 2-lead ECGs, the leads are not identical across datasets.

anyECG-Localization is further divided into two subsets: short-duration and long-duration. For the short-duration subset, ECGs are segmented into 10-second clips. For the long-duration subset, ECGs are segmented into clips of dynamic lengths ranging from 10 to 60 seconds. For each region where abnormalities occur, we resample 10 times for short-duration and 5 times for long-duration clips around the abnormal region, introducing a random time shift to enhance dataset diversity and robustness. To prevent the model from generating hallucinated responses (e.g., predicting abnormal regions when none exist), we included "Not Found" samples, where the queried feature is absent in the ECG. This ensures the model can correctly respond with "Not Found" instead of providing random time segments.

The dataset was reformatted into a QA structure. Questions are phrased as "Can you show me where the [abnormal] occurred on this ECG?" along with various paraphrased forms (see Table \ref{tab:template_localization}). Answers correspond to the localized waveform regions or "Not Found." Ultimately, anyECG-Localization comprises 100,050 short-duration ECG localization QA pairs and 50,025 long-duration ECG localization QA pairs. A portion of the dataset was reserved as a test set, ensuring that the same ECG (entire recording level, not segments level) does not appear in both the training and test sets.

\begin{tcolorbox}[colback=gray!10, colframe=black, boxrule=0.5pt, arc=3pt, width=\linewidth, title=Questions for Localization.]
  \begin{itemize}[leftmargin=0.5em]
    \scriptsize
      \item Can you show me where the \textcolor{blue}{\{abnormal\}} occurred on this ECG?
      \item Locate the \textcolor{blue}{\{abnormal\}} on this ECG for me, please.
      \item Could you identify where the \textcolor{blue}{\{abnormal\}} is on this ECG?
      \item Tell me where to find the \textcolor{blue}{\{abnormal\}} on this ECG.
      \item Please locate the specific location of the \textcolor{blue}{\{abnormal\}} on this ECG.
      \item Check this ECG and tell me where the \textcolor{blue}{\{abnormal\}} appears.
      \item Determine where the \textcolor{blue}{\{abnormal\}} is on this electrocardiogram.
      \item Help me find where the \textcolor{blue}{\{abnormal\}} shows up on this ECG.
      \item Examine this ECG and point out where the \textcolor{blue}{\{abnormal\}} is located.
      \item Assess this ECG and specify the location of the \textcolor{blue}{\{abnormal\}}.
      \item Where does the \textcolor{blue}{\{abnormal\}} appear in this ECG?
      \item On this ECG, where can I see the \textcolor{blue}{\{abnormal\}}?
      \item Can you locate the \textcolor{blue}{\{abnormal\}} on this ECG?
      \item Where is the \textcolor{blue}{\{abnormal\}} located in this ECG?
      \item Locate the \textcolor{blue}{\{abnormal\}} on this ECG for me, please.
      \item Could you point out where the \textcolor{blue}{\{abnormal\}} is on this ECG?
      \item Where should I look to find the \textcolor{blue}{\{abnormal\}} on this ECG?
      \item I need to find the \textcolor{blue}{\{abnormal\}} on this ECG; where should I look?
      \item Help me locate the \textcolor{blue}{\{abnormal\}} on this ECG.
      \item Determine where the \textcolor{blue}{\{abnormal\}} is located on this electrocardiogram.
  \end{itemize}
  \end{tcolorbox}
  \captionsetup{type=table}
  \captionof{table}{Questions for Localization.}
  \label{tab:template_localization}
 
\subsection{anyECG-MultiECG}
anyECG-MultiECG is a multi-ECG comparison dataset designed to address scenarios in clinical practice where physicians compare multiple ECGs from the same patient over time. It consists of two components: MIMIC Multi-ECG QA and ECG-QA \cite{oh2023ecg} constructed from PTB-XL \cite{wagner2020ptb}.

\begin{figure}[t]
  \centering
  \includegraphics[width=\linewidth]{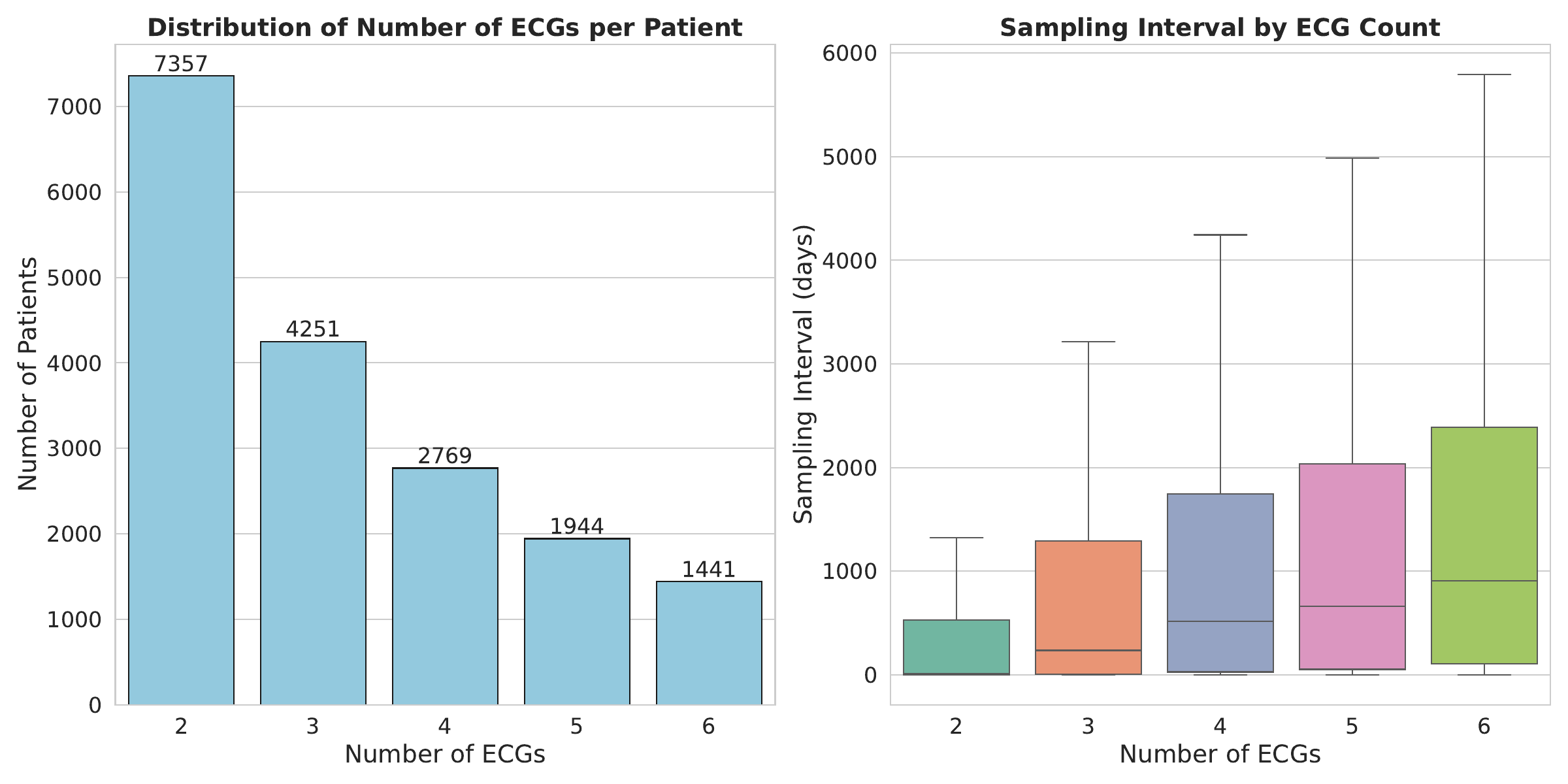}
  \caption{Statistics of the MIMIC Multi-ECG QA dataset.}
  \label{fig:mimic_statistic}
\end{figure}

The MIMIC Multi-ECG QA dataset is derived from the MIMIC-ECG dataset \cite{gowmimic}, which contains nearly 800,000 ECGs. Since the ability to compare multiple ECGs builds upon the model's understanding of single ECGs, only a small number of multi-ECG QA pairs are required for instruction tuning once the model has been trained on single ECG task. To construct this dataset, we selected the first 200,000 ECGs from MIMIC-ECG and grouped them by patient, identifying individuals with 2 to 6 ECGs. The distribution of the number of ECGs per patient and the sampling time intervals are detailed in Figure \ref{fig:mimic_statistic}. Open-ended QA pairs were generated using Llama-3.3-70B-Instruct \cite{touvron2023llama,grattafiori2024llama}, a pure language model. To provide the model with ECG information, we supplied the corresponding reports and the sampling times for each ECG. Six example questions were used as few-shot samples, covering various scenarios: (1) generating a report for each ECG, (2) providing a comprehensive diagnosis based on all ECGs, (3) identifying trends, and (4) predicting potential future changes. These four scenarios assume that the user provides only the order of the ECGs without specifying their sampling times. Additionally, we considered cases where sampling times are provided, including (5) absolute sampling times and (6) relative sampling times. The specific prompts are detailed in Table~\ref{tab:prompt_mimic}. For each patient, eight questions and corresponding answers were generated, resulting in a total of 135,094 multi-ECG QA pairs.

The second component, ECG-QA \cite{oh2023ecg}, is constructed from PTB-XL by \cite{wagner2020ptb}. Since the answers in ECG-QA are often overly simplistic (e.g., yes/no or a list of tags), we aimed to prevent the model from overfitting to this concise answering style. To achieve this, we used only one-tenth of the training set and appended the prompt "Please answer briefly." to the original questions. This subset contains 33,220 QA pairs.

\section{Evaluation Dataset Overview}
\label{sec:appendix_evaluation_dataset}
The evaluation datasets used to assess the performance of anyECG-chat are summarized in Table~\ref{tab:evaluation_dataset_statistic}. The datasets are categorized into three main tasks: ReportGen, Localization, and Multi-ECG.

\begin{table}[t]
  \centering
  \caption{Evaluation Dataset Overview}
  \resizebox{0.95\linewidth}{!}{%
  \begin{tabular}{@{}lll@{}}
    \toprule
    \textbf{Evaluation} & \textbf{Test QA} & \textbf{Setting} \\ 
    \midrule
    \multicolumn{3}{c}{\textbf{ReportGEN}} \\ \midrule
    PTBXL-Super & 2,158 & OOD \\ 
    PTBXL-Sub & 2,158 & OOD \\ 
    PTBXL-Form & 880 & OOD \\ 
    PTBXL-Rhythm & 2,098 & OOD \\ 
    CPSC & 1,382 & OOD \\ 
    CSN & 9,031 & OOD \\ 
    \midrule
    \multicolumn{3}{c}{\textbf{Localization}} \\ \midrule
    European ST-T Localization & 5,710 & 2 leads (ID), 1 lead (ZS) \\ 
    European ST-T Long Localization & 2,855 & 2 leads (ID), 1 lead (ZS) \\ 
    MIT-BIH ST Change Localization & 1,110 & 2 leads (ID), 1 lead (ZS) \\ 
    MIT-BIH ST Change Long Localization & 555 & 2 leads (ID), 1 lead (ZS) \\ 
    MIT-BIT Arrhythmia Localization & 10,230 & 2 leads (ID), 1 lead (ZS) \\ 
    MIT-BIT Arrhythmia Long Localization & 5,115 & 2 leads (ID), 1 lead (ZS) \\ 
    \midrule
    \multicolumn{3}{c}{\textbf{Multi-ECG}} \\ \midrule
    MIMIC Multi-ECG QA & 1152 & ID \\ 
    ECG-QA & 8,214 & 10\% train data used \\ 
    \bottomrule
  \end{tabular}%
  }
  \begin{tablenotes}
    \small
    \item \textbf{Notes:} ID: in-domain, OOD: out-of-domain, ZS: zero-shot. All datasets use the test subset.
  \end{tablenotes}
  \label{tab:evaluation_dataset_statistic}
\end{table}

\section{Human Scoring}
\begin{figure}[h]
  \centering
  \includegraphics[width=0.9\linewidth]{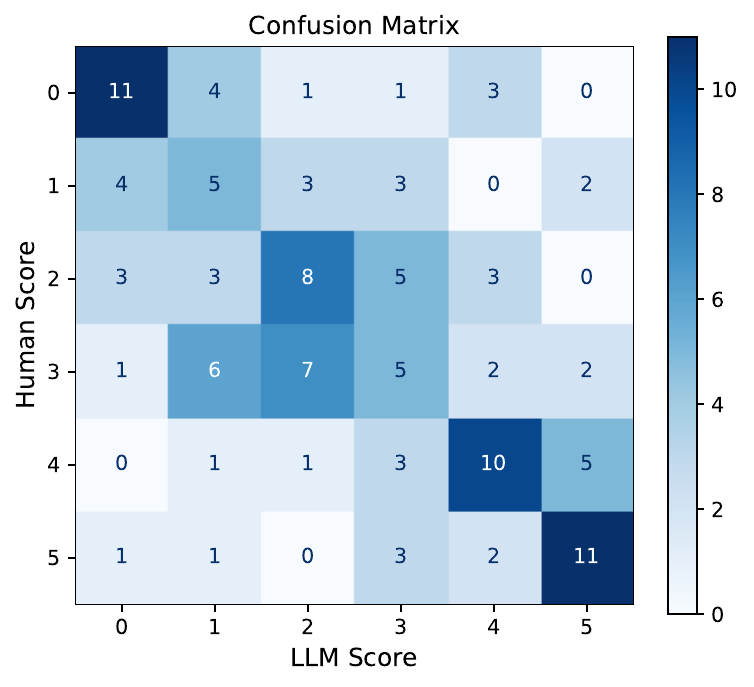}
  \caption{Comparison of LLM-based scoring and human scoring across different score levels.}
  \label{fig:human_scoring}
\end{figure}

To evaluate the reliability of LLM-based scoring in the MIMIC Multi-ECG QA assessment, we conducted a human scoring experiment. Specifically, we randomly selected 20 samples for each score level (0-5) assigned by the LLM and invited a medical student to perform manual scoring. The results, shown in Figure~\ref{fig:human_scoring}, demonstrate a strong consistency between the LLM and human evaluations.

\section{Case Study}

\subsection{Localization}
\label{sec:appendix_case_localization}
A case study of ECG localization is presented in Table~\ref{tab:case_localization}. As expected, LLaVa-Med and PULSE lack the fine-grained perceptual capability required for second-level abnormal waveform localization. When asked to identify the location of abnormal waveforms, they can only provide lead-level answers.

\begin{table}[h]
  \centering
  \caption{Case Study of ECG Localization.}
  \begin{tabular}{p{1.0\linewidth}}
  \toprule
  \textbf{ECG:} \\
  \includegraphics[width=0.7\linewidth]{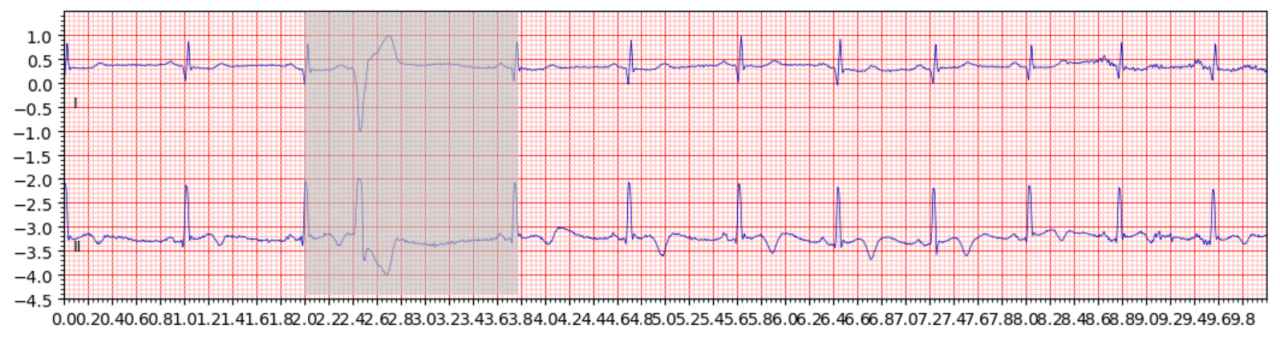} \\
  \textbf{Question:} Examine this ECG and point out where the Premature ventricular contraction is located. \\
  \textbf{Truth:} Duration: 2.0s-3.7s \\
  \midrule
  \textbf{anyECG-chat:} Duration: 1.9s-3.7s \\
  \textcolor{gray}{\textit{Others fail to localize at the second level.}} \\
  \textbf{LLaVa-Med:} The Premature ventricular contraction (PVC) is located in the V1-V2 region of the ECG. \\
  \textbf{MEIT:} V1-V2 \\
  \textbf{PULSE:} V2 \\
  \bottomrule
  \end{tabular}
  \label{tab:case_localization}
\end{table}

\begin{table*}[t]
\centering
\small
\begin{tcolorbox}[colback=gray!10, colframe=black, boxrule=0.5pt, arc=3pt, width=\textwidth, title=The prompt used to generate multi-ECG QA pairs.]
Based on the following ECGs, generate 8 different types of complex open-ended questions that require step-by-step thinking, and corresponding step-by-step answers. The following information is provided: the reports of each ECG and acquisition time. Questions should be about the ECG, in the question, you can choose to indicate the collection time of ECG or not. I need you to ask more questions. The more complex and diverse the question, the better. When the question q or answer a involves time, you need to provide the absolute or relative acquisition time of the ECG in the question.

For example, given reports:  
\textcolor{blue}{\texttt{[['Sinus tachycardia with PACs', 'Possible inferior infarct - age undetermined', 'Abnormal ECG'], ['Sinus arrhythmia'], ['Sinus rhythm', 'Probable left ventricular hypertrophy']]}  }
and acquisition time  
\textcolor{blue}{\texttt{['2148-11-12', '2149-06-06', '2149-12-24']}},  
\textcolor{blue}{\texttt{[0, 205, 406]}} days,
generate the following questions:  

\textcolor{mygreen}{\% ECG acquisition times are not provided, but the ECGs are presented in sequential order.}

\texttt{q: Provide a report for each electrocardiogram}  

\texttt{a: ECG1: Sinus tachycardia with PACs, possible inferior infarct - age undetermined, abnormal ECG. ECG2: Sinus arrhythmia. ECG3: Sinus rhythm, probable left ventricular hypertrophy.}  

\texttt{q: What can be found by combining these ECGs}  

\texttt{a: Combining these ECGs shows evolving cardiac patterns: initial tachycardia with possible infarct, followed by arrhythmia, then normalized rhythm with signs of left ventricular hypertrophy.}  

\texttt{q: What changes occur in the ECGs}  

\texttt{a: The ECGs show a shift from sinus tachycardia with PACs and possible infarct to sinus arrhythmia, then to normal sinus rhythm with probable left ventricular hypertrophy.}  

\texttt{q: Possible trends in the future}  

\texttt{a: Future ECGs may show progression of left ventricular hypertrophy or stabilization if underlying conditions are managed effectively.}  

\textcolor{mygreen}{\% Absolute Absolute ECG acquisition times are provided.}

\texttt{q: These electrocardiograms were taken on 2148-11-12, 2149-06-06, and 2149-12-24. Please help me take a look}  

\texttt{a: These ECGs from 2148-11-12 to 2149-12-24 show initial abnormalities, transient arrhythmia by mid-2149, and possible left ventricular hypertrophy by end of 2149.}  

\textcolor{mygreen}{\% Relative ECG acquisition times are provided.}

\texttt{q: The first ECG was collected 400 days ago, the second was collected 200 days ago, and the third was collected most recently. What changes have occurred?}  

\texttt{a: Over the past 400 days, ECGs show improvement from sinus tachycardia and possible infarct to normal rhythm, with recent signs of left ventricular hypertrophy.}  

Given \texttt{reports \textcolor{blue}{\{reports\}}}, and \texttt{acquisition time \textcolor{blue}{\{acquisition\_time\}}}, \textcolor{blue}{\texttt{\{acquisition\_time\_relative\}}}, generate 8 different types of complex open-ended questions that require step-by-step thinking, and corresponding step-by-step answers. Format each QA pair in a single line as a JSON dictionary (key “q” for question, and “a” for answer). Do not include any other explanation.
\end{tcolorbox}
\caption{The prompt used to generate multi-ECG QA pairs.}
\label{tab:prompt_mimic}
\end{table*}

\subsection{MIMIC Multi-ECG QA}
\label{sec:appendix_case_mimic_multi}
A detailed case study of MIMIC Multi-ECG QA is presented in Table~\ref{tab:case_mimic_multi}. We utilized QwQ-32B to assess the performance of anyECG-chat, LLaVa-Med, and PULSE. The evaluation prompt is provided at the bottom of Table~\ref{tab:case_mimic_multi}. To ensure fairness, we did not use the answers generated by Llama as the gold standard. Instead, we supplied the questions and corresponding reports as references, allowing QwQ to evaluate the quality of the model outputs based solely on this information.

\begin{table*}[h]
\centering
\caption{Case Study of MIMIC Multi-ECG QA.}
\begin{tabular}{@{}p{0.9\linewidth}@{}}
\toprule
\textbf{ECG:} \\
\includegraphics[width=0.7\linewidth]{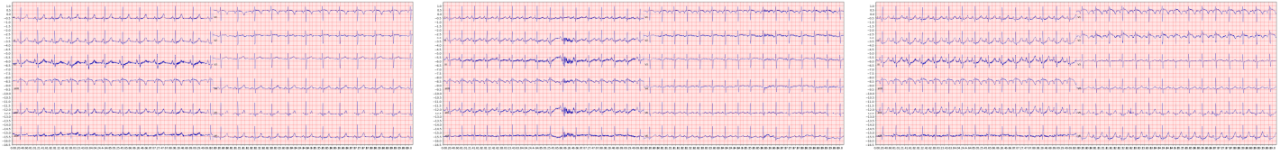} \\ 
\textbf{Reports from MIMIC-ECG:} \\
1. Sinus rhythm, Normal ECG \\
2. Sinus rhythm, Normal ECG \\
3. Sinus rhythm, Anterior T wave changes are nonspecific, Borderline ECG. \\
\textbf{Question:} Given the ECGs were collected 0, 716, and 1258 days apart, what can be inferred about the cardiac health over this period? \\
\midrule
\textbf{Answer (Llama) (given reports):} \\
Over 1258 days, the cardiac health appears to have remained largely stable, with sinus rhythm persisting, but with the emergence of nonspecific T wave changes, suggesting a need for continued monitoring. \\
\midrule
\textbf{Predictions (only ECG input)} \\
\textbf{anyECG-chat (Score 4):} \\
Over 1258 days, there was a progression from a normal ECG to a borderline ECG with possible left atrial abnormality and nonspecific T wave changes, suggesting a possible decline in cardiac health.\\
\textbf{LLaVa-Med (Score 2):} \\
It is difficult to draw specific conclusions about the cardiac health over this period without more information about the patient's medical history, symptoms, and any treatments or interventions they may have received during this time. However, the fact that the ECGs were collected at different time points can provide insights into the patient's cardiac health over time. By comparing the ECGs, healthcare professionals can assess changes in the patient's heart function, identify any potential issues, and monitor the effectiveness of any treatments or interventions. It is important to consult with a healthcare professional for a thorough evaluation and proper interpretation of the ECGs in the context of the patient's overall health. \\
\textbf{MEIT (Score 2):} \\
The ECGs show a persistent sinus rhythm over time, with later recordings indicating some nonspecific changes. \\
\textbf{PULSE (Score 1):} \\
Based on the ECGs, it can be inferred that there is no significant change in the cardiac health over this period, as the sinus tachycardia is consistent across all three ECGs. The presence of sinus tachycardia indicates that the heart is beating at an abnormally fast rate, but the underlying rhythm remains normal. The lack of any other abnormal findings or changes in the ECGs suggests that the cardiac health has not significantly deteriorated or improved over this period. \\
\midrule
\textbf{Prompt used for QwQ to score:} \\
\textcolor{mygreen}{\% Answer generated by Llama are not used as gold standard for evaluation.} \\
For the given question \textcolor{blue}{\{question\}} about multiple ECG-QA, and the report \textcolor{blue}{\{reports\}} corresponding to each ECG, score the answer below, where 0 means completely incorrect and 5 means completely correct. The answer is: \textcolor{blue}{<{prediction}>}. \\
\bottomrule
\end{tabular}
\label{tab:case_mimic_multi}
\end{table*}

\section{Multi-Turn QA}
\label{sec:appendix_case_multi_turn}
A case study of multi-turn QA is presented in Table~\ref{tab:case_multi_turn}. The example demonstrates how anyECG-chat can be used as a teaching tool for physicians, providing detailed explanations and suggestions based on the ECG data.

\begin{table*}[h]
\centering
\caption{Multi-Turn QA Example.}
\begin{tabular}{@{}p{0.9\linewidth}@{}}
\toprule
\textbf{ECG:} \\
\includegraphics[width=0.7\linewidth]{./pics/case_multi.png} \\ 
\midrule
\textbf{Turn 1:} \\
\textbf{User:} Given the ECGs were collected 0, 716, and 1258 days apart, what can be inferred about the cardiac health over this period? \\
\textbf{anyECG-chat:} Over 1258 days, there was a progression from a normal ECG to a borderline ECG with possible left atrial abnormality and nonspecific T wave changes, suggesting a possible decline in cardiac health. \\
\midrule
\textbf{Turn 2:} \\
\textbf{User:} What is the reason for the T wave change? \\
\textbf{anyECG-chat:} The T wave changes are nonspecific, meaning that they do not have a clear diagnostic significance. They could be due to a variety of factors, including electrolyte imbalance, cardiac ischemia, or other conditions. \\
\midrule
\textbf{Turn 3:} \\
\textbf{User:} If this patient is already 70 years old, based on the current ECG condition, do you have any suggestions? \\
\textbf{anyECG-chat:} Yes, considering the patient's age and the presence of nonspecific T wave changes, it's essential to monitor their cardiac health closely and consider preventive measures such as:

1. Regular follow-up appointments with a cardiologist.

2. Lifestyle modifications, such as a balanced diet, regular exercise, and stress management.

3. Control of underlying medical conditions that may contribute to cardiac health.

4. Consideration of additional diagnostic tests, such as echocardiography or stress testing, to further assess cardiac function and identify potential issues early.

Early detection and intervention can help prevent or mitigate the progression of cardiac disease in older adults.\\
\bottomrule
\end{tabular}
\label{tab:case_multi_turn}
\end{table*}

\end{document}